\documentclass[sigconf]{acmart}

\usepackage{booktabs} 
\usepackage{multirow}
\usepackage[acronym]{glossaries}
\usepackage{amsmath}
\usepackage{algorithm}
\usepackage{float}
\usepackage[noend]{algpseudocode}

\makeglossaries

\newacronym{GE}{GE}{Grammatical Evolution}
\newacronym{SGE}{SGE}{Structured Grammatical Evolution}
\newacronym{NN-SGE}{NN-SGE}{Neural Networks Structured Grammatical Evolution}
\newacronym{FFNN}{FFNN}{Feed-Forward Neural Network}
\newacronym{BNF}{BNF}{Backus-Naur Form}
\newacronym{GA}{GA}{Genetic Algorithm}
\newacronym{ANN}{ANN}{Artificial Neural Network}
\newacronym{GSNN}{GSNN}{Grammatical Swarm Neural Networks}
\newacronym{EANN}{EANN}{Evolutionary Artificial Neural Network}
\newacronym{GP}{GP}{Genetic Programming}
\newacronym{CFG}{CFG}{Context-Free Grammar}
\newacronym{EC}{EC}{Evolutionary Computation}
\newacronym{GGP}{GGP}{Grammar-based Genetic Programming}
\newacronym{RMSE}{RMSE}{Root Mean Squared Error}
\newacronym{EA}{EA}{Evolutionary Algorithms}
\newacronym{CNN}{CNN}{Convolutional Neural Network}
\newacronym{DSGE}{DSGE}{Dynamic Structured Grammatical Evolution}
\newacronym{ML}{ML}{Machine Learning}
\newacronym{DL}{DL}{Deep Learning}
\newacronym{AUROC}{AUROC}{Area Under the ROC Curve}
\newacronym{RPROP}{RPROP}{Resilient Backpropagation}

\setcopyright{rightsretained}

\copyrightyear{2017}
\acmYear{2017}
\setcopyright{acmcopyright}
\acmConference{GECCO '17}{July 15-19, 2017}{Berlin, Germany}\acmPrice{15.00}\acmDOI{http://dx.doi.org/10.1145/3071178.3071286} \acmISBN{978-1-4503-4920-8/17/07}

\begin{document}
\title{Towards the Evolution of Multi-Layered Neural Networks:\\A Dynamic Structured Grammatical Evolution Approach}

\author{Filipe Assun\c{c}\~ao, Nuno Louren\c{c}o, Penousal Machado, Bernardete Ribeiro}
\affiliation{CISUC, Department of Informatics Engineering,\\ University of Coimbra, Portugal}
\email{{fga, naml, machado, bribeiro}@dei.uc.pt}

\renewcommand{\shortauthors}{Filipe Assun\c{c}\~ao et al.}

\begin{abstract}
Current grammar-based NeuroEvolution approaches have several shortcomings. On the one hand, they do not allow the generation of \glspl{ANN} composed of more than one hidden-layer. On the other, there is no way to evolve networks with more than one output neuron. To properly evolve \glspl{ANN} with more than one hidden-layer and multiple output nodes there is the need to know the number of neurons available in previous layers. In this paper we introduce~\gls{DSGE}: a new genotypic representation that overcomes the aforementioned limitations. By enabling the creation of dynamic rules that specify the connection possibilities of each neuron, the methodology enables the evolution of multi-layered \glspl{ANN} with more than one output neuron. Results in different classification problems show that \gls{DSGE} evolves effective single and multi-layered \glspl{ANN}, with a varying number of output neurons.
\end{abstract}



\maketitle

\glsresetall

\section{Introduction}
\label{sec:intro}

\gls{ML} approaches, such as \glspl{ANN}, are often used to learn how to distinguish between multiple classes of a given problem. However, to reach near-optimal classifiers a laborious process of trial-and-error is needed to hand-craft and tune the parameters of \gls{ML} methodologies. In the specific case of \glspl{ANN} there are at least two manual steps that need to be considered: (i) the definition of the topology of the network, i.e., number of hidden-layers, number of neurons of each hidden-layer, and how should the layers be connected between each other; and (ii) the choice and parameterisation of the learning algorithm that is used to tune the weights and bias of the network connections (e.g., initial weights distribution and learning rate).

\glspl{EANN} or NeuroEvolution refers to methodologies that aim at the automatic search and optimisation of the \glspl{ANN}' parameters using \gls{EC}. With the popularisation of \gls{DL} and the need for \glspl{ANN} with a larger number of hidden-layers, NeuroEvolution has been vastly used in recent works \cite{tirumala2016evolving}.

The goal of the current work is the proposal of a novel \gls{GGP} methodology for evolving the topologies, weights and bias of \glspl{ANN}. We rely on a \gls{GGP} methodology because in this way we allow the direct encoding of different topologies for solving distinct problems in a plug-and-play fashion, requiring the user to set a grammar capable of describing the parameters of the \gls{ANN} to be optimised. One of the shortcomings of previous \gls{GGP} methodologies applied to NeuroEvolution is the fact that they are limited to the evolution of network topologies with one hidden-layer. The proposed approach, called \gls{DSGE} solves this constraint.

The remainder of the document is organised as follows: In Section~\ref{sec:background_sota} we detail \gls{SGE} and survey the state of the art on NeuroEvolution; Then, in Sections~\ref{sec:dsge} and~\ref{sec:multi_layer_nns}, \gls{DSGE} and its adaption to the evolution of multi-layered \glspl{ANN} are presented, respectively; The results and comparison of \gls{DSGE} with other \gls{GGP} approaches is conducted in Section~\ref{sec:experiments}; Finally, in Section~\ref{sec:conclusions}, conclusions are drawn and future work is addressed.

\section{Background and State of the Art}
\label{sec:background_sota}

In the following sub-sections we present \gls{SGE}, which serves as base for the development of \gls{DSGE}. Then, we survey \gls{EC} approaches for the evolution of \glspl{ANN}.

\subsection{Structured Grammatical Evolution}
\label{sec:sge}

\glsreset{SGE} \gls{SGE} was proposed by Louren\c{c}o et al.~\cite{lourenco2016unveiling} as a new genotypic representation for \gls{GE}~\cite{ryan1998ge}. The new representation aims at solving the redundancy and locality issues in \gls{GE}, and consists of a list of genes, one for each non-terminal symbol. Furthermore, a gene is a list of integers of the size of the maximum number of possible expansions for the non-terminal it encodes; each integer is a value in the interval $[0,\textit{non\_terminal\_possibilities}-1]$, where \textit{non\_terminal\_possibilities} is the number of possibilities for the expansion of the considered non-terminal symbol. Consequently, there is the need to pre-process the input grammar to find the maximum number of expansions for each non-terminal (further detailed in Section 3.1 of~\cite{lourenco2016unveiling}).

To deal with recursion, a set of intermediate symbols are created, which are no more than the same production rule replicated a pre-defined number of times (recursion level). The direct result of the \gls{SGE} representation is that, on the one hand, the decoding of individuals no longer depends on the modulus operation and thus its redundancy is removed; on the other hand, locality is enhanced, as codons are associated with specific non-terminals.

The genotype to phenotype mapping procedure is similar to \gls{GE}, with two main differences: (i) the modulus operation is not used; and (ii) integers are not read sequentially from a single list of integers, but are rather read sequentially from lists associated to each non-terminal symbol. An example of the mapping procedure is detailed in Section 3 of~\cite{lourenco2016unveiling}.
\subsection{NeuroEvolution}
\label{sec:sota}

Works focusing the automatic generation of \glspl{ANN} are often grouped according to the aspects of the \gls{ANN} they optimise: (i) evolution of the network parameters; (ii) evolution of the topology; and (iii) evolution of both the topology and parameters.


The gradient descent nature of learning algorithms, such as backpropagation, makes them likely to get trapped in local optima. One way to overcome this limitation is by using \gls{EC} to tune the \gls{ANN} weights and bias values. Two types of representation are commonly used: binary~\cite{whitley1990genetic} or real (e.g., CoSyNE~\cite{gomez2008accelerated}, Gravitational Swarm and Particle Swarm Optimization applied to OCR~\cite{fedorovici2013evolutionary}, training of deep neural networks \cite{david2014genetic}).

When training \glspl{ANN} using \gls{EC} the topology of the network is  provided and is kept fixed during evolution. Nevertheless, approaches tackling the automatic evolution of the topology of \glspl{ANN} have also been proposed. In this type of approaches for finding the adequate weights some authors use a off-the-shelf learning methodology (e.g., backpropagation) or evolve both the weights and the topology simultaneously.

Regarding the used representation for the evolution of the topology of \glspl{ANN} it is possible to divide the methodologies into two main types: those that use direct encodings (e.g., Topology-optimization Evolutionary Neural Network~\cite{rocha2007evolution}, NeuroEvolution of Augmenting Topologies~\cite{stanley2002evolving}) and those that use indirect encodings (e.g., Cellular Encoding~\cite{gruau1992genetic}). In the first two works the genotype is a direct representation of the network, and in the latter a mapping procedure has to be applied to transform the genotype into an interpretable network. Focusing on indirect representations, in the next section we further detail grammar-based approaches.

\subsection{Grammar-based NeuroEvolution}
\label{sec:sota_grammars}

Over the last years several approaches applying \gls{GE} to NeuroEvolution have been proposed. However, the works that focus the evolution of the networks topology are limited to generating one-hidden-layered \glspl{ANN}, because of the difficulties in tracking the number of neurons available in previous layers.

Tsoulos et al.~\cite{tsoulos2008neural} and Ahmadizar et al.~\cite{ahmadizar2015artificial} describe two different approaches based on \gls{GE} for the evolution of both the topology and parameters of one-hidden-layered \glspl{ANN}. While the first evolves the topology and weights using \gls{GE}, the latter combines \gls{GE} with a \gls{GA}: \gls{GE} is applied to the evolution of the topology and the \gls{GA} is used for searching the real values (i.e., weights and bias). The use of a \gls{GA} to optimise the real values is motivated by the fact that \gls{GE} is not suited for the evolution and tuning of real values. For that reason, Soltanian et al.~\cite{soltanian2013artificial} just use \gls{GE} to optimise the topology of the network, and rely on the backpropagation algorithm to train the evolved topologies.

Although \gls{GE} is the most common approach for the evolution of \glspl{ANN} by means of grammars it is not the only one. Si et al. in~\cite{si2014grammatical} present \gls{GSNN}: an approach that uses Grammatical Swarm for the evolution of the weights and bias of a fixed \gls{ANN} topology, and thus, \gls{EC} is used just for the generation of real numbers. In~\cite{jung2006evolutionary}, Jung and Reggia detail a method for searching adequate topologies of \glspl{ANN} based on descriptive encoding languages: a formal way of defining the environmental space and how should individuals be formed; the \gls{RPROP} algorithm is used for training the \glspl{ANN}. \gls{SGE} has also been used for the  evolution of the topology and weights of~\glspl{ANN}~\cite{previouswork}.



\section{Dynamic Structured Grammatical Evolution}
\label{sec:dsge}

\glsreset{DSGE}\gls{DSGE} is our novel \gls{GGP} approach. With the proposed methodology the gain is twofold: (i) all the genotype is used, i.e., while in \gls{GE} and \gls{SGE} the genotype encodes the largest allowed sequence, in \gls{DSGE} the genotype grows as needed; and (ii) there is no need to pre-process the grammar in order to compute the maximum tree-sizes of each non-terminal symbol, so that intermediate grammar derivation rules are created. In the next sections we describe in detail the components that make these gains possible.

\subsection{Representation}
\label{sec:representation}

\begin{figure}
\centering
\footnotesize
\begin{tabbing}
\hspace{1cm}~$<$start$>$ \= $::=$ \= $<$float$>$~~~~~ \= (0) \\
\end{tabbing}\vspace{-0.3cm}\begin{tabbing}
\hspace{1cm}~$<$float$>$ \= $::=$ \= $<$first$>$.$<$second$>$~~~~ \= (0) \\
\end{tabbing}\vspace{-0.3cm}\begin{tabbing}
\hspace{1cm}~$<$first$>$ \= $::=$ \= $0$~~~ \= (0)\\
 ~~~~~~~ \> $|$ \> $1$~~ \> (1)\\
 ~~~~~~~ \> $|$ \> $2$~~ \> (2)\\
\end{tabbing}\vspace{-0.3cm}\begin{tabbing}
\hspace{1cm}~$<$second$>$ \= $::=$ \= $<$digit$>$ $<$second$>$~~~~~~~~~~~ \= (0) \\
 ~~~~~~~ \> $|$ \> $<$digit$>$~~~~~~~~~~~ \> (1)\\
\end{tabbing}\vspace{-0.3cm}\begin{tabbing}
\hspace{1cm}~$<$digit$>$ \= $::=$ \= $0$ ~~~~~~~~~~~ \= ~~~~ (0)\\
 ~~~~~~~ \> $|$ \> $\ldots$~~~~~~~~~~~ \= ~~($\ldots$)\\
 ~~~~~~~ \> $|$ \> $9$ ~~~~~~~~~~~ \> ~~(9)\\
\end{tabbing}

\includegraphics[width=0.35\textwidth]{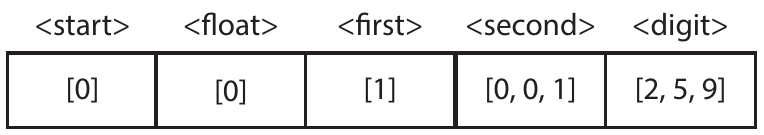}
\caption{Example of a grammar (top) and of a genotype of a candidate solution (bottom).}
\label{fig:representation}
\end{figure}

Each candidate solution encodes an ordered sequence of the derivation steps of the used grammar that are needed to generate a specific solution for the problem at hand. The representation is similar to the one used in \gls{SGE}, with one main difference: instead of computing and generating the maximum number of derivations for each of the grammar's non-terminal symbols, a variable length representation is used, where just the number of needed derivations are encoded. Consequently, there is no need to create intermediate symbols to deal with recursive rules. 

To limit the genotype size, a maximum depth value is defined for each non-terminal symbol. Allowing different limits for each non-terminal symbol provides an intuitive and flexible way of constraining the search space by limiting, for instance, the maximum number of hidden-layers, neurons or connections.

Figure~\ref{fig:representation} represents an example grammar for the generation of real numbers in the $[0, 3[$ interval, along with the representation of a candidate solution encoding the number $1.259$. The genotype is encoded as a list of genes, where each gene encodes an ordered sequence of choices for expanding a given non-terminal symbol, as a list of integers. The genotype to phenotype mapping is detailed in Section~\ref{sec:mapping}.

\subsection{Initialisation}
\label{sec:initialisation}

\begin{algorithm}[t!]
\caption{Random candidate solution creation.}\label{alg:initialisation}
\footnotesize
\begin{algorithmic}[1]
\Procedure{create\_individual}{grammar, max\_depth, genotype, symbol, depth}

    \State expansion = randint(0, len(grammar[symbol])-1)
    \If{is\_recursive(symbol)}
        \If{expansion in grammar.recursive(symbol)}
            \If{depth $\ge$ max\_depth[symbol]}:
                \State non\_rec = grammar.non\_recursive(symbol)
                \State expansion = choice(non\_rec)
            \EndIf
        \EndIf
    \EndIf
    
    \If{symbol in genotype}
        \State{genotype[symbol].append(expansion)}
    \Else
        \State{genotype[symbol] = [expansion]}
    \EndIf
    
    \State expansion\_symbols = grammar[symbol][expansion]
    
    \For{sym in expansion\_symbols}
        \If{not is\_terminal(sym)}
            \If{symbol == sym}
                \State create\_individual(grammar, max\_depth, genotype, symbol, depth+1)
            \Else
                \State create\_individual(grammar, max\_depth, genotype, symbol, 0)
            \EndIf
        \EndIf
    \EndFor
\EndProcedure
\end{algorithmic}
\end{algorithm}

Algorithm~\ref{alg:initialisation} details the recursive function that is used to generate each candidate solution. The input parameters are: the grammar that describes the domain of the problem; the maximum depth of each non-terminal symbol; the genotype (which is initially empty); the non-terminal symbol that we want to expand (initially the start symbol is used); and the current sub-tree depth (initialised to $0$). Then, for the non-terminal symbol given as input, one of the possible derivation rules is selected (lines 2-11) and the non-terminal symbols of the chosen derivation rule are recursively expanded (lines 12-18). However, when selecting the expansion rule there is the need to check whether or not the maximum sub-tree depth has already been reached (lines 3-5). If that happens, only non-recursive derivation rules can be selected for expanding the current non-terminal symbol (lines 6-7). This procedure is repeated until an initial population with the desired size is created.

\subsection{Mapping Function}
\label{sec:mapping}

To map the candidate solutions genotype into the phenotype that will later be interpreted as an \gls{ANN} we use Algorithm~\ref{alg:mapping}. The algorithm is similar to the one used to generate the initial population but, instead of randomly selecting the derivation rule to use in the expansion of the non-terminal symbol, we use the choice that is encoded in the individual's genotype (lines 12-22). During evolution the genetic operators may change the genotype in a way that requires a larger number of integers than the ones available. When this happens, the following genotype's repair procedure is applied: new derivation rules are selected at random and added to the genotype of the individual (lines 3-11). In addition to returning the genotype the algorithm also computes which genotype integers are being used, which will later help in the application of the genetic operators. Table~\ref{tab:mapping_example} shows an example of the mapping procedure applied to the genotype of Figure~\ref{fig:representation}. 

\begin{algorithm}[t!]
\caption{Genotype to phenotype mapping procedure.}\label{alg:mapping}
\footnotesize
\begin{algorithmic}[1]
\Procedure{mapping}{genotype, grammar, max\_depth, read\_integers, symbol, depth}

    \State phenotype = ``''

    \If{symbol not in read\_integers}
        \State read\_integers[symbol] = 0
    \EndIf
    \If{symbol not in genotype}
        \State genotype[symbol] = []
    \EndIf

    \If{read\_integers[symbol] $\ge$ len(genotype[symbol])}
        \If{depth $\ge$ max\_depth[symbol]}
            \State generate\_terminal\_expansion(genotype, symbol)
        \Else
            \State generate\_expansion(genotype, symbol)
        \EndIf
    \EndIf
    
    \State gen\_int = genotype[symbol][read\_integer[symbol]]
    \State expansion = grammar[symbol][gen\_int]
    \State read\_integers[symbol] += 1
    
    \For{sym in expansion}
        \If{is\_terminal(sym)}
            \State phenotype += sym
        \Else
            \If{symbol == sym}
                \State phenotype += mapping(genotype, grammar, max\_depth, read\_integers, sym, depth+1)
            \Else
                \State phenotype += mapping(genotype, grammar, max\_depth, read\_integers, sym, 0)
            \EndIf
        \EndIf
    \EndFor
    
    \State \textbf{return} phenotype

\EndProcedure
\end{algorithmic}
\end{algorithm}

\subsection{Genetic Operators}
\label{sec:genetic_operators}

To explore the problem's domain and therefore promote evolution we rely on mutation and crossover. 

\paragraph{Mutation} is restricted to integers that are used in the genotype to phenotype mapping and changes a randomly selected expansion option (encoded as an integer) to another valid one, constraint to the restrictions on the maximum sub-tree depth. To do so, we first select one gene; the probability of selecting the $i$-th gene ($p_i$) is proportional to the number of integers of that non-terminal symbol that are used in the genotype to phenotype mapping ($read\_integers$):
\begin{equation*}
    p_i = \frac{read\_integers_i}{\sum_{j=1}^n read\_integers_j},
\end{equation*}
where $n$ is the total number of genes. Additionally, genes where there is just one possibility for expansion (e.g, $<$start$>$ or $<$float$>$ of the grammar of Figure~\ref{fig:representation}) are not considered for mutation purposes. After selecting the gene to be mutated, we randomly select one of its integers and replace it with another valid possibility. 

Considering the genotype of Figure~\ref{fig:representation}, a possible result from the application of the mutation operator is $[[0], [0], [2], [0,0,1], [2,5,9]]$, that represents the number $2.259$.

\paragraph{Crossover} is used to recombine two parents (selected using tournament selection) to generate two offspring. We use one-point crossover. As such, after selecting the cutting point (at random) the genetic material is exchanged between the two parents. The choice of the cutting point is done at the gene level and not at the integers level, i.e., what is exchanged between parents are genes and not the integers.

Given two parents [[0], [0], [2], $|$ [0,0,1], [2,5,9]] and [[0], [0], [1], $|$ [0,0,0,1], [1,0,2,4]], where $|$ denotes the cutting point, the generated offspring would be  [[0], [0], [2], [0,0,0,1], [1,0,2,4]] and [[0], [0], [1], [0,0,1], [2,5,9]].

\begin{table}[t!]
    \centering
    \caption{Example of the mapping procedure. Each row represents a derivation step. The list of genes represents the integers needed for expanding start, float, first, second and digit, respectively.}
    \label{tab:mapping_example}
    \footnotesize
    \begin{tabular}{c|l}
        Derivation step & Integers left  \\ \hline 
        $<$start$>$ & $[[0], [0], [1], [0,0,1], [2,5,9]]$\\ 
        $<$float$>$ & $[[], [0], [1], [0,0,1], [2,5,9]]$ \\              
        $<$first$>$.$<$second$>$ & $[[], [], [1], [0,0,1], [2,5,9]]$ \\        
        1.$<$second$>$  & $[[], [], [], [0,0,1], [2,5,9]]$\\      
        1.$<$digit$><$second$>$ &  $[[], [], [], [0,1], [2,5,9]]$\\
        1.2$<$second$>$ &  $[[], [], [], [0,1], [5,9]]$\\      
        1.2$<$digit$><$second$>$ & $[[], [], [], [1], [5,9]]$\\
        1.25$<$second$>$ &  $[[], [], [], [1], [9]]$\\       
        1.25$<$digit$>$  & $[[], [], [], [], [9]]$\\                
        1.259 & $[[], [], [], [], []]$
    \end{tabular}
\end{table}

\subsection{Fitness Evaluation}


To enable the comparison of the evolved \glspl{ANN} with the results from a previous work~\cite{previouswork} the performance is measured as the \gls{RMSE} obtained while solving a classification task. In addition, for better dealing with unbalanced datasets, this metric considers the \gls{RMSE} per class, and the resulting fitness function is the multiplication of the exponential values of the multiple \glspl{RMSE} per class, as follows:
\begin{equation*}
\text{fitness} = \prod_{c=1}^{m} exp\Bigg(\sqrt{\frac{\sum_{i=1}^{n_c} (o_i - t_i)^2 }{n_c}} \, \Bigg),
\end{equation*}
where $m$ is the number of classes of the problem, $n_c$ is the number of instances of the problem that belong to class $c$, $o_i$ is the confidence value predicted by the evolved network, and $t_i$ is the target value. This way, higher errors are more penalised than lower ones, helping the evolved networks to better generalise to unseen data.

\section{Evolution of Multi-Layered ANN\small{s}}
\label{sec:multi_layer_nns}

Figure~\ref{fig:one_layer_grammar} shows the grammar that was used in~\cite{previouswork}, which is similar to those used in other works focusing the grammar-based evolution of \glspl{ANN}~\cite{tsoulos2008neural,soltanian2013artificial,ahmadizar2015artificial}. The rationale behind the design of this grammar is the evolution of networks composed of one hidden-layer, where only the hidden-neurons as well as the weights of the connections from the input and to the output neurons are evolved. 

Three major drawbacks can be pointed to the previous grammatical formulation: (i) it only allows the generation of networks with one hidden-layer; (ii) the output neuron is always connected to all neurons in the hidden-layer; and (iii) there is no way to define multiple output nodes, at least one that reuses the hidden-nodes, instead of creating new ones.




These drawbacks are related to limitations of the evolutionary engines that are overcome by the approach presented in this paper.
Figure~\ref{fig:multi_layer_grammar} represents a grammar capable of representing multi-layered \glspl{ANN}, with a variable number of neurons in each layer. However, there is no way of knowing how many neurons are in the previous layer, so that the established connections are valid. In the next sections we detail the adaptions we introduce to allow the evolution of multi-layered networks.

\subsection{Dynamic Production Rules}
\label{sec:dynamic_rules}

To know the neurons available in each hidden-layer we create new production rules, in run-time. More precisely, considering the grammar of Figure~\ref{fig:multi_layer_grammar}, for each $i$-th $<$layer$>$ non-terminal symbol we create a $<$features-i$>$ production rule, encoding the features that layer can use as input.

\begin{figure}[t!]
    \centering
    \footnotesize
    \begin{tabbing}
\hspace{1cm}~$<$sigexpr$>$ \= $::=$ \= $<$node$>$ \\
 ~~~~~~~ \> $|$ \>  $<$node$>$ $+$ $<$sigexpr$>$\\
\end{tabbing}\vspace{-0.3cm}\begin{tabbing}
\hspace{1cm}~$<$node$>$ \= $::=$ \= $<$weight$>$ $*$ sig($<$sum$>$ $+$ $<$bias$>$) \\
\end{tabbing}\vspace{-0.3cm}\begin{tabbing}
\hspace{1cm}~$<$sum$>$ \= $::=$ \= $<$weight$>$ $*$ $<$features$>$\\
 ~~~~~~~ \> $|$ \> $<$sum$>$ + $<$sum$>$  \\
\end{tabbing}\vspace{-0.3cm}\begin{tabbing}
\hspace{1cm}~$<$features$>$ \= $::=$ \= $x_1$ \\
 ~~~~~~~ \> $|$ \> $\ldots$\\
 ~~~~~~~ \> $|$ \> $x_n$ \\
\end{tabbing}\vspace{-0.3cm}\begin{tabbing}
\hspace{1cm}~$<$weight$>$ \= $::=$ \= $<$number$>$\\
\end{tabbing}\vspace{-0.3cm}\begin{tabbing}
\hspace{1cm}~$<$bias$>$ \= $::=$ \= $<$number$>$\\
\end{tabbing}\vspace{-0.3cm}\begin{tabbing}
\hspace{1cm}~$<$number$>$ \= $::=$ \= $<$digit$>$.$<$digit$>$$<$digit$>$ \\
 ~~~~~~~ \> $|$ \> --$<$digit$>$.$<$digit$>$$<$digit$>$\\
\end{tabbing}\vspace{-0.3cm}\begin{tabbing}
\hspace{1cm}~$<$digit$>$ \= $::=$ \= $0$  $|$ $1$ $|$ $2$ $|$ $3$ $|$ $4$  \\
 ~~~~~~~ \> $|$ \> $5$ $|$ $6$ $|$ $7$ $|$ $8$ $|$ $9$\\
\end{tabbing}
    \caption{Grammar used in~\cite{previouswork}. $n$ represents the number of features of the problem.}
    \label{fig:one_layer_grammar}
\end{figure}

For the first hidden-layer, $<$features-1$>$ has the available features as expansion possibilities, i.e., the ones that are initially defined in the grammar ($x_1$, $\dots$, $x_n$, where $n$ represents the number of features). Then, for the next hidden-layers there are two possibilities: (i) let the connections be established to all the neurons in previous layers (including input neurons); or (ii) limit the connections to the neurons in the previous layer. 

In the current work we have decided for the first option, with the restriction that the neurons in the output layer can only be connected to hidden-nodes. When establishing the connections between neurons, the probability of choosing a neuron in the previous layer is proportional to the number of previous hidden-layers. More specifically:
\begin{equation*}
    P(neuron_{i-1}) = \sum_{j=1}^{j=i-2} P(neuron_j),
\end{equation*}
i.e., when establishing the connections of the $i$-th layer, the probability of linking to a neuron in the previous layer ($i-1$) is equal to the probability of linking to a neuron in the remaining layers ($1$, $\ldots$, $i-2$). The rationale is to minimise the emergence of deep networks with useless neurons, in the sense that they are not connected (directly or indirectly) to output nodes.


\begin{figure}[t!]
    \centering
    \footnotesize
    \begin{tabbing}
\hspace{1cm}~$<$start$>$ \= $::=$ \= $<$hidden-layers$>$ \texttt{--} $<$output-layer$>$ \\
\end{tabbing}\vspace{-0.3cm}\begin{tabbing}
\hspace{1cm}~$<$hidden-layers$>$ \= $::=$ \= $<$hidden-layers$>$ \texttt{--} $<$hidden-layers$>$ \\
~~~~~~~ \> $|$ \> $<$layer$>$ \texttt{--} $<$hidden-layers$>$ \\
~~~~~~~ \> $|$ \> $<$layer$>$  \\
\end{tabbing}\vspace{-0.3cm}\begin{tabbing}
\hspace{1cm}~$<$output-layer$>$ \= $::=$ \= sig( $<$sum$>$ + $<$float$>$) \\
\end{tabbing}\vspace{-0.3cm}\begin{tabbing}
\hspace{1cm}~$<$layer$>$ \= $::=$ \= $<$nodes$>$ \\
\end{tabbing}\vspace{-0.3cm}\begin{tabbing}
\hspace{1cm}~$<$nodes$>$ \= $::=$ \= $<$nodes$>$ \texttt{-} $<$nodes$>$ \\
~~~~~~~ \> $|$ \> $<$node$>$ \texttt{-} $<$nodes$>$ \\
~~~~~~~ \> $|$ \> $<$node$>$ \texttt{-} $<$node$>$  \\
\end{tabbing}\vspace{-0.3cm}\begin{tabbing}
\hspace{1cm}~$<$node$>$ \= $::=$ \= sig($<$sum$>$ $+$ $<$float$>$) \\
\end{tabbing}\vspace{-0.3cm}\begin{tabbing}
\hspace{1cm}~$<$sum$>$ \= $::=$ \= $<$float$>$ $*$ $<$features$>$\\
 ~~~~~~~ \> $|$ \> $<$sum$>$ + $<$sum$>$  \\
 ~~~~~~~ \> $|$ \> $<$sum$>$ + $<$sum$>$  \\
\end{tabbing}\vspace{-0.3cm}
\begin{tabbing}
\hspace{1cm}~$<$float$>$ \= $::=$ \= $<$digit$>$.$<$digit$>$$<$digit$>$ \\
 ~~~~~~~ \> $|$ \> --$<$digit$>$.$<$digit$>$$<$digit$>$\\
\end{tabbing}\vspace{-0.3cm}\begin{tabbing}
\hspace{1cm}~$<$digit$>$ \= $::=$ \= $0$  $|$ $1$ $|$ $2$ $|$ $3$ $|$ $4$  \\
 ~~~~~~~ \> $|$ \> $5$ $|$ $6$ $|$ $7$ $|$ $8$ $|$ $9$\\
\end{tabbing}\vspace{-0.3cm}\begin{tabbing}
\hspace{1cm}~$<$features$>$ \= $::=$ \= $x_1$ \\
 ~~~~~~~ \> $|$ \> $\ldots$\\
 ~~~~~~~ \> $|$ \> $x_n$ \\
\end{tabbing}
    \vspace{-10pt}
    \caption{Grammar used for evolving multi-layered \glspl{ANN}. $n$ represents the number of features of the problem; \texttt{-} and \texttt{--} are used as neuron and layer separators, respectively.}
    \label{fig:multi_layer_grammar}
\end{figure}

For example, using the grammar of Figure~\ref{fig:multi_layer_grammar}, when generating an \gls{ANN} with two hidden-layers, with 4 and 3 nodes, respectively, the following production rules are created and added to the grammar:
\begin{itemize}
\item[] $<$features-1$>$ $::=$ $x_1$ $|$ $\ldots$ $|$ $x_n$
\item[] $<$features-2$>$ $::=$ $h_{1,1}$ $|$ $h_{1,2}$ $|$ $h_{1,3}$ $|$ $h_{1,4}$
\item[] $<$features-3$>$ $::=$ $h_{2,1}$ $|$ $h_{2,2}$ $|$ $h_{2,3}$,
\end{itemize}
where $x$ represents input features, $n$ the total number of input features, and $h_{i,j}$ the output of the $j$-th neuron of the $i$-th layer.

\subsection{Evolutionary Engine Components}
\label{sec:evo_engine_components}

To cope with dynamic derivation rules some of the evolutionary components  detailed in Section~\ref{sec:dsge} are adapted. The representation of the candidate solutions is kept the same, with the additional genes for each of the dynamic rules (e.g., $<$\mbox{features-i}$>$) that are created. The number of dynamic rules can be different from individual to individual, making the grammar a property of the individual.

When mapping the genotype to the phenotype, if during the expansion of a non-terminal symbol a dynamic rule is called, it is necessary to know  which one of the possibilities should be used. For example, in the example provided in the end of Section~\ref{sec:dynamic_rules}, when expanding the $<$sum$>$ non-terminal symbol we need to know to which dynamic rule $<$features$>$ corresponds to: $<$features-1$>$, $<$features-2$>$ or $<$\mbox{features-3}$>$. Thus, we add a $num\_layer$ parameter to Algorithm~\ref{alg:mapping}, which is initialised at zero and incremented by one each time a new $<$layer$>$ non-terminal symbol is expanded. Then, when the $<$features$>$ non-terminal symbol is read, we expand it based on the derivation rule $<$\mbox{features-num\_layer}$>$. However, before applying the expansion possibility it is necessary to check if it is valid. The genetic operators may have changed the number of neurons in one of the layers, or even the number of existing layers. Consequently, when the mapping procedure is applied,  the genotype may be encoding a connection to a neuron in a previous layer that has been erased. In such a scenario, the genotype is fixed by replacing the faulty connection with a valid one. The dynamic rule is also corrected.

Finally, it is important to mention that now the number of genes can vary from individual to individual, which is generated by a potentially different number of hidden-layers. Therefore, in the application of the crossover operator, we only allow the exchange of the genes that are common to both parents.

\section{Experimental Results}
\label{sec:experiments}

The conducted experiments are divided into two steps. First, we use the same grammar as in~\cite{previouswork} to check whether or not \gls{DSGE} is capable of generating one-hidden-layered \glspl{ANN} that perform better than those found using~\gls{GE} and~\gls{SGE}. Then, using a new grammatical formulation, we check whether or not \gls{DSGE} is suitable for the evolution of multi-layered \glspl{ANN}.

\subsection{Datasets}
\label{sec:datasets}

\begin{table}[t!]
\centering
\footnotesize
\caption{Properties of the used datasets.}
\label{tab:dataset_properties}
\begin{tabular}{c|c|c|c|c|}
\cline{2-5}
                                     & Flame   & WDBC    & Ionosphere & Sonar   \\ \hline
\multicolumn{1}{|c|}{Num. Features}  & 2       & 30      & 34         & 60      \\ \hline
\multicolumn{1}{|c|}{Num. Instances} & 240     & 569     & 351        & 208     \\ \hline
\multicolumn{1}{|c|}{Class 0}        & 36.25\% & 62.74\% & 35.90\%    & 53.37\% \\ \hline
\multicolumn{1}{|c|}{Class 1}        & 63.75\% & 37.26\% & 64.10\%    & 46.63\% \\ \hline
\end{tabular}
\end{table}

We selected four binary classification problems. The problems have an increasing complexity in terms of the available number of features of the classification task that is to be performed (see Table~\ref{tab:dataset_properties}). In the next paragraphs we present a brief description of the datasets. We focus on binary problems to enable comparison with other grammar-based approaches.

\begin{description}
    \item[Flame~\cite{fu2007flame} --] This dataset contains artificially generated data for clustering purposes. 
    \item[Wisconsin Breast Cancer Detection (WDBC)~\cite{street1993nuclear,uclrepository} --] The WDBC dataset is comprised of features extracted from digitalised images of breast masses that are to be classified into malign and benign. 
     \item[Ionosphere~\cite{sigillito1989classification,uclrepository} --] This benchmark is used for the classification of ionosphere radar returns, where the returns are classified into two different classes: good if it returns evidences of structure; and bad otherwise.
    \item[Sonar~\cite{uclrepository,gorman1988analysis} --] The sonar dataset contains features extracted from sonar signals that allow a classification model to separate between signals that are bounced off a metal cylinder or a rock cylinder. 
\end{description}

\subsection{Grammar}

Two grammars are used in the conducted experiments. For the first set of experiments, targeting the evolution of one-hidden-layered \glspl{ANN} the same grammar of~\cite{previouswork} is used (see Figure ~\ref{fig:one_layer_grammar}). This grammar allows the evolution of the topology and parameters of \glspl{ANN}, where all the hidden-nodes are connected to the output-node; the evolved topologies cannot have more than one hidden-layer and are restricted to just one output neuron.

Then we introduce the grammar detailed in Figure~\ref{fig:multi_layer_grammar}, which enables the generation of non fully-connected layers with more than one hidden-layer and a varying number of neurons in each layer. Note that it is not required that all neurons are directly or indirectly connected to the output nodes. Also, recall from Section~\ref{sec:dynamic_rules} that the connections can be established to any input or hidden-node in previous layers, except for the connections of the output nodes that can only be established to hidden-nodes. In the production rules $<$hidden-layers$>$, $<$nodes$>$ and $<$sum$>$ a higher probability is given to the recursive expansion of the non-terminal symbol, due to the higher difficulty of tuning parameters in deeper and more complex topologies.

\setlength{\tabcolsep}{1pt}
\begin{table}[t!]
    \centering
    \footnotesize
    \caption{Experimental parameters.}
    \label{tab:setup}
    \begin{tabular}{c|c}
        \textbf{Parameter} & \textbf{Value} \\ \hline
        Num. runs & 30 \\
        Population size & 100 \\
        Num. generations (1 hidden-layer) & 500 \\
        Num. generations ($\ge1$ hidden layers) & 1500  / 3500\\
        Crossover rate & 95\% \\
        Mutation rate & 1 mutation in 95\% of the individuals \\
        Tournament size & 3 \\
        Elite size & 1\% \\ 
        \textbf{GE Parameter} & \textbf{Value} \\ \hline
        Individual size & 200 \\
        Wrapping & 0 \\
        \textbf{SGE Parameter} & \textbf{Value} \\ \hline
        Recursion level & 6 \\ 
        \textbf{DSGE Parameter} & \textbf{Value} \\ \hline
        Max. depth (1 hidden-layer) &  \{sigexpr: 6, sum: 3\} \\
        Max. depth ($\ge1$ hidden-layers) &  \{hidden-layers: 3, nodes: 5, sum: 4\} \\
        \textbf{Dataset Parameter} & \textbf{Value} \\ \hline
        Training percentage & 70\% \\
        Testing percentage & 30\% \\
    \end{tabular}
\end{table}

\subsection{Experimental Setup}

Table~\ref{tab:setup} details the experimental parameters used for the tests conducted with \gls{GE}, \gls{SGE} and \gls{DSGE}. The parameters where $1$ hidden-layer and $\ge 1$ hidden-layers are mentioned refer to the experiments conducted with the grammars of Figures~\ref{fig:one_layer_grammar} and \ref{fig:multi_layer_grammar}, respectively. 

To make the exploration of the search space similar we only allow one mutation in 95\% of the population individuals. In the search for multi-layered networks the domain is bigger and, consequently we perform longer runs. For the experiments focusing on the generation of one-hidden-layered networks we restrict the domains to allowing similar networks in terms of the neurons and connections they can have: \gls{GE}, \gls{SGE} and one-hidden-layered \gls{DSGE} are constrained to generated networks with up to $7$ neurons and $8$ connections in each neuron. In the experiments targeting networks that can have more than one hidden-layer these limits are increased: the maximum number of neurons in each layer is $32$, the maximum number of connections of each neuron is $16$; it is possible to generate networks with up to $8$ hidden-layers.


The used datasets are all randomly partitioned in the same way: $70\%$ of each class instances are used for training and the remaining $30\%$ for testing. We only measure the individuals performance using the train data, and thus the test data is kept aside the evolutionary process and exclusively used for validation purposes, i.e., to evaluate the behaviour of the networks in the classification of instances that have not been seen during the creation of the \gls{ANN}. The datasets are used as they are obtained, i.e., no pre-processing or data augmentation methodologies are applied.

\subsection{Evolution of One-Hidden-Layered ANNs}

\begin{figure}[t!]
\centering
\includegraphics[width=0.4\textwidth]{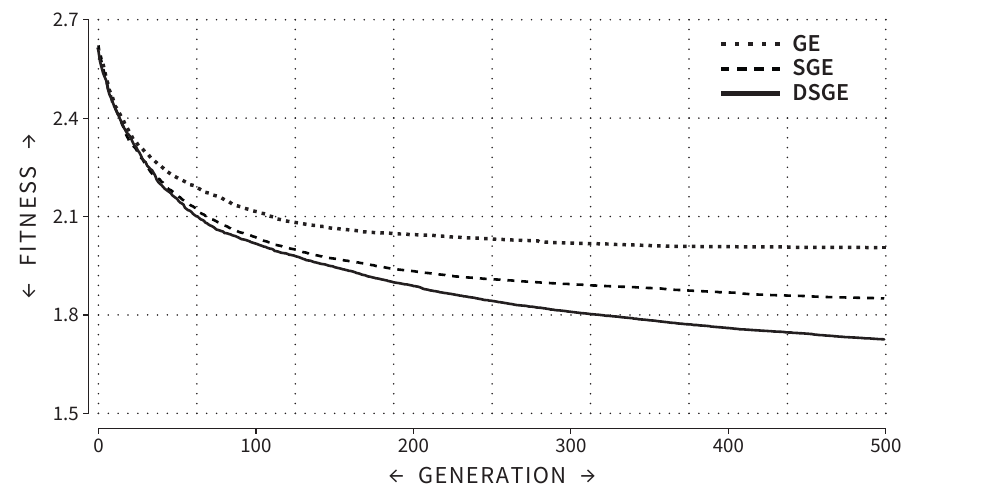}
\caption{Fitness evolution of the best individuals across generations for the sonar dataset.}
\label{fig:fitness_evolution}
\end{figure}

To investigate whether or not \gls{DSGE} performs better than previous grammar-based approaches we focus on experiments for the evolution of one-hidden-layered \glspl{ANN}. For that, we use the grammar of Figure~\ref{fig:one_layer_grammar}. 

Figure~\ref{fig:fitness_evolution} shows the evolution of the fitness across generations for \gls{GE}, \gls{SGE} and \gls{DSGE} on the sonar dataset. Because of space constraints we only present the fitness evolution results on the most challenging dataset. Results are averages of $30$ independent runs. The figure clearly shows \gls{DSGE} outperforms the other methods, indicating that the new representation, where no mutation is silent promotes locality and the efficient exploration of the search space.

Table~\ref{tab:one_layer_results} reports the results obtained with \gls{DSGE}, which are compared with other grammar-based approaches: \gls{GE} and \gls{SGE}. Results are averages of the best network (in terms of fitness) of each of the evolutionary runs, and are formatted as follows: mean $\pm$ standard deviation. For all experiments we record fitness, \gls{RMSE}, \gls{AUROC} and f-measure. Except for the fitness metric, which is only computed over the train set, all metrics are calculated for the train and test sets. In addition, we also report the average number of neurons and used features. 


An analysis of the results shows that \gls{DSGE} performs better than the other approaches for all the datasets, and for all the metrics considered. 
Regarding the structure of the generated networks, \gls{DSGE} is capable of finding solutions which are more complex in terms of the number of used neurons and input features. As the complexity of the networks grows more real values have to be tuned; nonetheless, \gls{DSGE} is capable of performing such tuning, reaching solutions that, despite being more complex, perform better in the tested classification problems.

\setlength{\tabcolsep}{0.8pt}
\begin{table}[t!]
\footnotesize
\caption{Evolution of one-hidden-layered \glspl{ANN}. Comparison between \gls{DSGE} and other grammar-based approaches. Results are averages of $30$ independent runs. $+$ and $\sim$ represent the result of statistical tests (see text).}
\label{tab:one_layer_results}
\begin{tabular}{ccc|c|c|c|c|}
\cline{4-7}
                                              &                                                 &                           & Flame                 & WDBC                  & Ionosphere            & Sonar               \\ \hline
\multicolumn{2}{|c|}{\multirow{3}{*}{Fitness}}                                                  & GE                        & 1.58 $\pm$ 0.36         & 1.55$\pm$ 0.18          & 1.82 $\pm$ 0.28         & 2.01 $\pm$ 0.23             \\
\multicolumn{2}{|c|}{}                                                                          & SGE                       & 1.32 $\pm$ 0.25         & 1.46 $\pm$ 0.08         & 1.48 $\pm$ 0.18         & 1.85 $\pm$ 0.18                \\
\multicolumn{2}{|c|}{}                                                                          & \multicolumn{1}{c|}{DSGE} & \multicolumn{1}{c|}{1.16 $\pm$ 0.13$^{+++}$} & \multicolumn{1}{c|}{1.36 $\pm$ 0.06$^{+++}$} & \multicolumn{1}{c|}{1.38 $\pm$ 0.13$^{+++}$} & \multicolumn{1}{c|}{1.73 $\pm$ 0.12$^{+++}$}  \\ \hline
\multicolumn{1}{|c|}{\multirow{12}{*}{\rotatebox[origin=c]{90}{Train}}} & \multicolumn{1}{c|}{\multirow{3}{*}{RMSE}}      & GE                        & 0.28 $\pm$ 0.15         & 0.24 $\pm$ 0.09         & 0.33 $\pm$ 0.10         & 0.38 $\pm$ 0.08                  \\
\multicolumn{1}{|c|}{}                        & \multicolumn{1}{c|}{}                           & SGE                       & 0.16 $\pm$ 0.13         & 0.19 $\pm$ 0.03         & 0.21 $\pm$ 0.07         & 0.34 $\pm$ 0.06                \\
\multicolumn{1}{|c|}{}                        & \multicolumn{1}{c|}{}                           & \multicolumn{1}{c|}{DSGE} & \multicolumn{1}{c|}{0.08 $\pm$ 0.06$^{+++}$} & \multicolumn{1}{c|}{0.15 $\pm$ 0.03$^{+++}$} & \multicolumn{1}{c|}{0.17 $\pm$ 0.05$^{++}$} & \multicolumn{1}{c|}{0.30 $\pm$ 0.05$^{++}$} \\ \cline{2-7} 
\multicolumn{1}{|c|}{}                        & \multicolumn{1}{c|}{\multirow{3}{*}{Accuracy}}  & GE                        & 0.90 $\pm$ 0.10         & 0.92 $\pm$ 0.12          & 0.79 $\pm$ 0.20         & 0.76 $\pm$ 0.16                 \\
\multicolumn{1}{|c|}{}                        & \multicolumn{1}{c|}{}                           & SGE                       & 0.96 $\pm$ 0.08         & 0.95 $\pm$ 0.02         & 0.93 $\pm$ 0.11         & 0.84 $\pm$ 0.12                \\
\multicolumn{1}{|c|}{}                        & \multicolumn{1}{c|}{}                           & \multicolumn{1}{c|}{DSGE} & \multicolumn{1}{c|}{0.99 $\pm$ 0.03$^{+++}$} & \multicolumn{1}{c|}{0.97 $\pm$ 0.01$^{+++}$} & \multicolumn{1}{c|}{0.97 $\pm$ 0.03$^{+++}$} & \multicolumn{1}{c|}{0.90 $\pm$ 0.04$^{++}$} \\ \cline{2-7} 
\multicolumn{1}{|c|}{}                        & \multicolumn{1}{c|}{\multirow{3}{*}{AUROC}}     & GE                        & 0.96 $\pm$ 0.05         & 0.98 $\pm$ 0.02         & 0.86 $\pm$ 0.18         & 0.88 $\pm$ 0.06                 \\
\multicolumn{1}{|c|}{}                        & \multicolumn{1}{c|}{}                           & SGE                       & 0.98 $\pm$ 0.04         & 0.99 $\pm$ 0.01         & 0.94 $\pm$ 0.04         & 0.91 $\pm$ 0.04                \\
\multicolumn{1}{|c|}{}                        & \multicolumn{1}{c|}{}                           & \multicolumn{1}{c|}{DSGE} & \multicolumn{1}{c|}{0.99 $\pm$ 0.01$^{+++}$} & \multicolumn{1}{c|}{0.99 $\pm$ 0.00$^{+++}$} & \multicolumn{1}{c|}{0.96 $\pm$ 0.03$^{+++}$} & \multicolumn{1}{c|}{0.93 $\pm$ 0.04$^{\sim}$}  \\ \cline{2-7} 
\multicolumn{1}{|c|}{}                        & \multicolumn{1}{c|}{\multirow{3}{*}{F-measure}} & GE                        & 0.91 $\pm$ 0.09         & 0.88 $\pm$ 0.18         & 0.78 $\pm$ 0.32         & 0.70 $\pm$ 0.29                 \\
\multicolumn{1}{|c|}{}                        & \multicolumn{1}{c|}{}                           & SGE                       & 0.96 $\pm$ 0.08         & 0.93 $\pm$ 0.03         & 0.93 $\pm$ 0.18         & 0.78 $\pm$ 0.23                \\
\multicolumn{1}{|c|}{}                        & \multicolumn{1}{c|}{}                           & \multicolumn{1}{c|}{DSGE} & \multicolumn{1}{c|}{0.99 $\pm$ 0.02$^{+++}$} & \multicolumn{1}{c|}{0.96 $\pm$ 0.02$^{+++}$} & \multicolumn{1}{c|}{0.98 $\pm$ 0.02$^{+++}$} & \multicolumn{1}{c|}{0.88 $\pm$ 0.05$^{++}$} \\ \hline
\multicolumn{1}{|c|}{\multirow{12}{*}{\rotatebox[origin=c]{90}{Test}}}  & \multicolumn{1}{c|}{\multirow{3}{*}{RMSE}}      & GE                        & 0.31 $\pm$ 0.15         & 0.27 $\pm$ 0.09         & 0.38 $\pm$ 0.07         & 0.45 $\pm$ 0.06                  \\
\multicolumn{1}{|c|}{}                        & \multicolumn{1}{c|}{}                           & SGE                       & 0.22 $\pm$ 0.13         & 0.23 $\pm$ 0.04         & 0.32 $\pm$ 0.05         & 0.44 $\pm$ 0.04                \\
\multicolumn{1}{|c|}{}                        & \multicolumn{1}{c|}{}                           & \multicolumn{1}{c|}{DSGE} & \multicolumn{1}{c|}{0.14 $\pm$ 0.08$^{++}$} & \multicolumn{1}{c|}{0.20 $\pm$ 0.03$^{++}$} & \multicolumn{1}{c|}{0.28 $\pm$ 0.04$^{+++}$} & \multicolumn{1}{c|}{0.43 $\pm$ 0.04$^{\sim}$}  \\ \cline{2-7} 
\multicolumn{1}{|c|}{}                        & \multicolumn{1}{c|}{\multirow{3}{*}{Accuracy}}  & GE                        & 0.88 $\pm$ 0.11         & 0.90 $\pm$ 0.12         & 0.76 $\pm$ 0.18         & 0.68 $\pm$ 0.12                  \\
\multicolumn{1}{|c|}{}                        & \multicolumn{1}{c|}{}                           & SGE                       & 0.93 $\pm$ 0.09         & 0.93 $\pm$ 0.02         & 0.87 $\pm$ 0.10         & 0.73 $\pm$ 0.09                \\
\multicolumn{1}{|c|}{}                        & \multicolumn{1}{c|}{}                           & \multicolumn{1}{c|}{DSGE} & \multicolumn{1}{c|}{0.97 $\pm$ 0.05$^{+++}$} & \multicolumn{1}{c|}{0.95 $\pm$ 0.02$^{+++}$} & \multicolumn{1}{c|}{0.90 $\pm$ 0.03$^{++}$} & \multicolumn{1}{c|}{0.76 $\pm$ 0.05$^{\sim}$}  \\ \cline{2-7} 
\multicolumn{1}{|c|}{}                        & \multicolumn{1}{c|}{\multirow{3}{*}{AUROC}}     & GE                        & 0.93 $\pm$ 0.08         & 0.97 $\pm$ 0.03         & 0.83 $\pm$ 0.09         & 0.81 $\pm$ 0.05                  \\
\multicolumn{1}{|c|}{}                        & \multicolumn{1}{c|}{}                           & SGE                       & 0.96 $\pm$ 0.08         & 0.98 $\pm$ 0.02         & 0.90 $\pm$ 0.05         & 0.82 $\pm$ 0.05                 \\
\multicolumn{1}{|c|}{}                        & \multicolumn{1}{c|}{}                           & \multicolumn{1}{c|}{DSGE} & \multicolumn{1}{c|}{0.99 $\pm$ 0.03$^{++}$} & \multicolumn{1}{c|}{0.98 $\pm$ 0.01$^{++}$} & \multicolumn{1}{c|}{0.93 $\pm$ 0.04$^{++}$} & \multicolumn{1}{c|}{0.83 $\pm$ 0.04$^{\sim}$}  \\ \cline{2-7} 
\multicolumn{1}{|c|}{}                        & \multicolumn{1}{c|}{\multirow{3}{*}{F-measure}} & GE                        & 0.89 $\pm$ 0.09         & 0.86 $\pm$ 0.18         & 0.76 $\pm$ 0.31         & 0.61 $\pm$ 0.25                 \\
\multicolumn{1}{|c|}{}                        & \multicolumn{1}{c|}{}                           & SGE                       & 0.94 $\pm$ 0.09         & 0.91 $\pm$ 0.03         & 0.89 $\pm$ 0.17         & 0.64 $\pm$ 0.20                \\
\multicolumn{1}{|c|}{}                        & \multicolumn{1}{c|}{}                           & \multicolumn{1}{c|}{DSGE} & \multicolumn{1}{c|}{0.98 $\pm$ 0.04$^{+++}$} & \multicolumn{1}{c|}{0.93 $\pm$ 0.03$^{+++}$} & \multicolumn{1}{c|}{0.93 $\pm$ 0.02$^{++}$} & \multicolumn{1}{c|}{0.72 $\pm$ 0.06$^{\sim}$}  \\ \hline
\multicolumn{2}{|c|}{\multirow{3}{*}{Num. Neurons}}                                             & GE                        & 3.33 $\pm$ 1.40         & 3.13 $\pm$ 1.53         & 2.50 $\pm$ 1.41         & 2.53 $\pm$ 1.20               \\
\multicolumn{2}{|c|}{}                                                                          & SGE                       & 4.87 $\pm$ 1.83         & 3.73 $\pm$ 1.53         & 3.53 $\pm$ 1.36         & 3.07 $\pm$ 1.39                 \\
\multicolumn{2}{|c|}{}                                                                          & \multicolumn{1}{c|}{DSGE} & \multicolumn{1}{c|}{6.47 $\pm$ 1.20} & \multicolumn{1}{c|}{6.23 $\pm$ 1.58} & \multicolumn{1}{c|}{5.97 $\pm$ 1.78} & \multicolumn{1}{c|}{6.13 $\pm$ 1.69} \\ \hline
\multicolumn{2}{|c|}{\multirow{3}{*}{Num. Features}}                                            & GE                        & 1.97 $\pm$ 0.18         & 8.40 $\pm$ 3.81         & 7.33 $\pm$ 5.33         & 9.40 $\pm$ 5.73                 \\
\multicolumn{2}{|c|}{}                                                                          & SGE                       & 2.00 $\pm$ 0.00         & 12.0 $\pm$ 6.51         & 12.1 $\pm$ 5.79         & 13.3 $\pm$ 6.42                \\
\multicolumn{2}{|c|}{}                                                                          & \multicolumn{1}{c|}{DSGE} & \multicolumn{1}{c|}{2.00 $\pm$ 0.00} & \multicolumn{1}{c|}{14.5 $\pm$ 3.52} & \multicolumn{1}{c|}{13.3 $\pm$ 4.74} & \multicolumn{1}{c|}{17.6 $\pm$ 4.66} \\ \hline
\end{tabular}
\end{table}

To verify if the differences between the tested approaches are significant we perform a statistical analysis. In~\cite{previouswork} we have already demonstrated that \gls{SGE} is consistently statistically superior to \gls{GE}. As such, we will now focus in analysing if \gls{DSGE} is statistically superior to \gls{SGE}. To check if the samples follow a Normal Distribution we use the Kolmogorov-Smirnov and Shapiro-Wilk tests, with a significance level of $\alpha = 0.05$. The tests revealed that data does not follow any distribution and, as such, a non-parametric test (Mann-Whitney U, $\alpha$ = 0.05) will be used to perform the pairwise comparison for each recorded metric. Table \ref{tab:one_layer_results} uses a graphical overview to present the results of the statistical analysis: $\sim$ indicates no statistical difference between \gls{DSGE} and \gls{SGE} and $+$ signals that \gls{DSGE} is statistically superior to \gls{SGE}. The effect size is denoted by the number of $+$ signals, where $+$, $++$ and $+++$ correspond respectively to low (0.1 $\leq$ r $<$ 0.3), medium (0.3 $\leq$ r $<$ 0.5) and large (r $\geq$ 0.5) effect sizes. The statistical analysis reveals that \gls{DSGE} is consistently statistically superior to \gls{SGE}: \gls{DSGE} is never worse than \gls{SGE} and is only equivalent in $5$ situations. In all the remaining 31 comparisons, \gls{DSGE} is statistically superior, with a medium effect size in $11$ occasions and a large effect size in $20$ occasions.


Focusing on the comparison between  train and test set results in \gls{DSGE} the differences between train and test performance are, on average, $0.09$, $0.06$, $0.04$ and $0.06$, for  \gls{RMSE}, accuracy, \gls{AUROC} and f-measure, respectively.  For \gls{SGE} the differences are $0.08$, $0.06$, $0.04$ and $0.06$ and for \gls{GE} $0.05$, $0.04$, $0.04$ and $0.04$, for the \gls{RMSE}, accuracy, \gls{AUROC} and f-measure, respectively. Despite the fact that the differences in \gls{DSGE} are superior to the ones in \gls{SGE} and \gls{GE}, it is our perception that this is a result of the better \gls{DSGE} results and not an indicator of overfitting. 

In~\cite{previouswork} we showed that the results obtained with \gls{SGE} are superior to those of other grammar-based approaches (namely, the ones described in~\cite{tsoulos2008neural,ahmadizar2015artificial,soltanian2013artificial}). Additionally, we showed that it is beneficial to evolve both the topology and weights of a network, since it has better results than \glspl{ANN} obtained by hand-crafting the topology of the networks and training them using backpropagation. As~\gls{DSGE} is statistically superior to~\gls{SGE} it is then clear that it is also better than previous methods. 


\subsection{Evolution of Multi-Layered ANNs}

\begin{table}[t!]
\centering
\footnotesize
\caption{\gls{DSGE} evolution of multi-layered \glspl{ANN}. $+$ and $\sim$ symbols represent the result of statsitical tests, and have the same meaning as in Table~\ref{tab:one_layer_results}.}
\label{tab:results_multilayer}
\begin{tabular}{ccc|c|c|c|c|}
\cline{4-7}
                                             &                                                 &            & Flame         & WDBC          & Ionosphere    & Sonar         \\ \hline
\multicolumn{2}{|c|}{\multirow{3}{*}{Fitness}}                                                 & 1 H-L & 1.16 $\pm$ 0.13 & 1.36 $\pm$ 0.06 & 1.38 $\pm$ 0.13 & 1.73 $\pm$ 0.12 \\
\multicolumn{2}{|c|}{}                                                                         & $\ge1$ H-Ls & 1.15 $\pm$ 0.18$^{\sim}$ & 1.35 $\pm$ 0.13$^{\sim}$ & 1.36 $\pm$ 0.19$^{\sim}$ & 1.57 $\pm$ 0.18$^{\sim}$ \\
\multicolumn{2}{|c|}{}                                                                         & $>1$ H-Ls & 1.08 $\pm$ 0.13$^{+++}$ & 1.32 $\pm$ 0.11$^{\sim}$ & 1.30 $\pm$ 0.11$^{+++}$ & 1.53 $\pm$ 0.15$^{+++}$ \\ \hline

\multicolumn{1}{|c|}{\multirow{12}{*}{\rotatebox[origin=c]{90}{Train}}} & \multicolumn{1}{c|}{\multirow{3}{*}{RMSE}}      & 1 H-L & 0.08 $\pm$ 0.06 & 0.15 $\pm$ 0.03 & 0.17 $\pm$ 0.05 & 0.30 $\pm$ 0.05 \\ 
\multicolumn{1}{|c|}{}                       & \multicolumn{1}{c|}{}                           & $\ge1$ H-Ls & 0.07 $\pm$ 0.08$^{\sim}$ & 0.15 $\pm$ 0.05$^{\sim}$ & 0.17 $\pm$ 0.08$^{\sim}$ & 0.26 $\pm$ 0.07$^{\sim}$ \\  
\multicolumn{1}{|c|}{}                       & \multicolumn{1}{c|}{}                           & $>1$ H-Ls & 0.07 $\pm$ 0.05$^{+++}$ & 0.07 $\pm$ 0.04$^{\sim}$ & 0.14 $\pm$ 0.05$^{+++}$ & 0.25 $\pm$ 0.06$^{+++}$ \\ \cline{2-7} 

\multicolumn{1}{|c|}{}                       & \multicolumn{1}{c|}{\multirow{3}{*}{Accuracy}}  & 1 H-L & 0.99 $\pm$ 0.03 & 0.97 $\pm$ 0.01 & 0.97 $\pm$ 0.03 & 0.90 $\pm$ 0.04 \\
\multicolumn{1}{|c|}{}                       & \multicolumn{1}{c|}{}                           & $\ge1$ H-Ls & 0.99 $\pm$ 0.02$^{\sim}$ &0.97 $\pm$ 0.02$^{\sim}$ & 0.97 $\pm$ 0.03$^{\sim}$ & 0.92 $\pm$ 0.05$^{\sim}$ \\ 
\multicolumn{1}{|c|}{}                       & \multicolumn{1}{c|}{}                           & $>1$ H-Ls & 0.99 $\pm$ 0.02$^{\sim}$ & 0.97 $\pm$ 0.02$^{\sim}$ & 0.98 $\pm$ 0.02$^{\sim}$ & 0.93 $\pm$ 0.04$^{+++}$ \\ \cline{2-7} 

\multicolumn{1}{|c|}{}                       & \multicolumn{1}{c|}{\multirow{3}{*}{AUROC}}     & 1 H-L & 0.99 $\pm$ 0.01 & 0.99 $\pm$ 0.00 & 0.96 $\pm$ 0.03 & 0.93 $\pm$ 0.04 \\
\multicolumn{1}{|c|}{}                       & \multicolumn{1}{c|}{}                           & $\ge1$ H-Ls & 0.99 $\pm$ 0.01$^{\sim}$ & 0.99 $\pm$ 0.01$^{\sim}$ & 0.96 $\pm$ 0.04$^{\sim}$ & 0.93 $\pm$ 0.05$^{\sim}$ \\ 
\multicolumn{1}{|c|}{}                       & \multicolumn{1}{c|}{}                           & $>1$ H-Ls & 1.00 $\pm$ 0.01$^{\sim}$ & 0.99 $\pm$ 0.00$^{\sim}$ & 0.97 $\pm$ 0.02$^{\sim}$ & 0.93 $\pm$ 0.04$^{\sim}$ \\ \cline{2-7} 

\multicolumn{1}{|c|}{}                       & \multicolumn{1}{c|}{\multirow{3}{*}{F-measure}} & 1 H-L & 0.99 $\pm$ 0.02 & 0.96 $\pm$ 0.02 & 0.98 $\pm$ 0.02 & 0.88 $\pm$ 0.05 \\
\multicolumn{1}{|c|}{}                       & \multicolumn{1}{c|}{}                           & $\ge1$ H-Ls & 0.99 $\pm$ 0.02$^{\sim}$ & 0.96 $\pm$ 0.03$^{\sim}$ & 0.97 $\pm$ 0.02$^{\sim}$ & 0.91 $\pm$ 0.06$^{\sim}$ \\ 
\multicolumn{1}{|c|}{}                       & \multicolumn{1}{c|}{}                           & $>1$ H-Ls & 0.99 $\pm$ 0.01$^{\sim}$ & 0.97 $\pm$ 0.03$^{\sim}$ & 0.98 $\pm$ 0.01$^{+++}$ & 0.92 $\pm$ 0.05$^{+++}$ \\ \hline

\multicolumn{1}{|c|}{\multirow{12}{*}{\rotatebox[origin=c]{90}{Test}}}  & \multicolumn{1}{c|}{\multirow{3}{*}{RMSE}}      & 1 H-L & 0.14 $\pm$ 0.08 & 0.20 $\pm$ 0.03 & 0.28 $\pm$ 0.04 & 0.43 $\pm$ 0.04 \\
\multicolumn{1}{|c|}{}                       & \multicolumn{1}{c|}{}                           & $\ge1$ H-Ls & 0.17 $\pm$ 0.08$^{\sim}$ & 0.20 $\pm$ 0.04$^{\sim}$ & 0.31 $\pm$ 0.05$^{\sim}$ & 0.44 $\pm$ 0.05$^{\sim}$ \\ 
\multicolumn{1}{|c|}{}                       & \multicolumn{1}{c|}{}                           & $>1$ H-Ls & 0.16 $\pm$ 0.08$^{\sim}$ & 0.20 $\pm$ 0.04$^{\sim}$ & 0.29 $\pm$ 0.05$^{\sim}$ & 0.43 $\pm$ 0.06$^{\sim}$ \\ \cline{2-7} 

\multicolumn{1}{|c|}{}                       & \multicolumn{1}{c|}{\multirow{3}{*}{Accuracy}}  & 1 H-L & 0.97 $\pm$ 0.05 & 0.95 $\pm$ 0.02 & 0.90 $\pm$ 0.03 & 0.76 $\pm$ 0.05 \\
\multicolumn{1}{|c|}{}                       & \multicolumn{1}{c|}{}                           & $\ge1$ H-Ls & 0.96 $\pm$ 0.04$^{\sim}$ & 0.95 $\pm$ 0.02$^{\sim}$ & 0.89 $\pm$ 0.03$^{\sim}$ & 0.77 $\pm$ 0.05$^{\sim}$ \\ 
\multicolumn{1}{|c|}{}                       & \multicolumn{1}{c|}{}                           & $>1$ H-Ls & 0.97 $\pm$ 0.04$^{\sim}$ & 0.95 $\pm$ 0.02$^{\sim}$ & 0.90 $\pm$ 0.03$^{\sim}$ & 0.78 $\pm$ 0.06$^{\sim}$ \\ \cline{2-7} 

\multicolumn{1}{|c|}{}                       & \multicolumn{1}{c|}{\multirow{3}{*}{AUROC}}     & 1 H-L & 0.99 $\pm$ 0.03 & 0.98 $\pm$ 0.01 & 0.93 $\pm$ 0.04 & 0.83 $\pm$ 0.04 \\
\multicolumn{1}{|c|}{}                       & \multicolumn{1}{c|}{}                           & $\ge1$ H-Ls & 0.98 $\pm$ 0.03$^{\sim}$ & 0.99 $\pm$ 0.01$^{\sim}$ & 0.91 $\pm$ 0.06$^{\sim}$ & 0.82 $\pm$ 0.07$^{\sim}$ \\ 
\multicolumn{1}{|c|}{}                       & \multicolumn{1}{c|}{}                           & $>1$ H-Ls & 0.99 $\pm$ 0.02$^{\sim}$ & 0.99 $\pm$ 0.01$^{\sim}$ & 0.93 $\pm$ 0.03$^{\sim}$ & 0.82 $\pm$ 0.07$^{\sim}$ \\ \cline{2-7} 

\multicolumn{1}{|c|}{}                       & \multicolumn{1}{c|}{\multirow{3}{*}{F-measure}} & 1 H-L & 0.98 $\pm$ 0.04 & 0.93 $\pm$ 0.03 & 0.93 $\pm$ 0.02 & 0.72 $\pm$ 0.06 \\
\multicolumn{1}{|c|}{}                       & \multicolumn{1}{c|}{}                           & $\ge1$ H-Ls & 0.97 $\pm$ 0.03$^{\sim}$ & 0.93 $\pm$ 0.03$^{\sim}$ & 0.92 $\pm$ 0.02$^{\sim}$ & 0.73 $\pm$ 0.07$^{\sim}$ \\ 
\multicolumn{1}{|c|}{}                       & \multicolumn{1}{c|}{}                           & $>1$ H-Ls & 0.97 $\pm$ 0.03$^{\sim}$ & 0.93 $\pm$ 0.03$^{\sim}$ & 0.92 $\pm$ 0.02$^{\sim}$ & 0.74 $\pm$ 0.08$^{\sim}$ \\ \hline

\multicolumn{2}{|c|}{\multirow{3}{*}{Num. H-Ls}}                                             & 1 H-L & 1.00 $\pm$ 0.00 & 1.00 $\pm$ 0.00 & 1.00 $\pm$ 0.00 & 1.00 $\pm$ 0.00 \\
\multicolumn{2}{|c|}{}                                                                         & $\ge1$ H-Ls & 2.27 $\pm$ 0.98 & 2.23 $\pm$ 1.04 & 1.90 $\pm$ 0.84 & 2.37 $\pm$ 0.96 \\ 
\multicolumn{2}{|c|}{}                                                                         & $>1$ H-Ls & 2.81 $\pm$ 0.60 & 2.95 $\pm$ 0.52 & 2.50 $\pm$ 0.51 & 2.95 $\pm$ 1.12 \\ \hline

\multicolumn{2}{|c|}{\multirow{3}{*}{Num. Neurons}}                                            & 1 H-L & 3.33 $\pm$ 1.40 & 3.13 $\pm$ 1.53 & 2.50 $\pm$ 1.41 & 2.53 $\pm$ 0.38  \\
\multicolumn{2}{|c|}{}                                                                         & $\ge1$ H-Ls & 10.6 $\pm$ 6.49 & 9.73 $\pm$ 7.03 & 7.17 $\pm$ 4.98 & 9.27 $\pm$ 4.55 \\ 
\multicolumn{2}{|c|}{}                                                                         & $>1$ H-Ls & 12.5 $\pm$ 6.64 & 12.8 $\pm$ 6.70 & 9.33 $\pm$ 5.24 & 10.9 $\pm$ 4.12 \\ \hline

\multicolumn{2}{|c|}{\multirow{3}{*}{Num. Features}}                                           & 1 H-L & 2.00 $\pm$ 0.00 & 14.5 $\pm$ 3.52 & 13.1 $\pm$ 6.87 & 19.8 $\pm$ 8.39 \\
\multicolumn{2}{|c|}{}                                                                         & $\ge1$ H-Ls & 2.00 $\pm$ 0.00 & 16.3 $\pm$ 7.03 & 13.8 $\pm$ 6.66 & 21.8 $\pm$ 8.48 \\ 
\multicolumn{2}{|c|}{}                                                                         & $>1$ H-Ls & 2.00 $\pm$ 0.00 & 18.2 $\pm$ 6.91 & 15.8 $\pm$ 5.48 & 24.0 $\pm$ 7.54 \\ \hline

\end{tabular}
\end{table}



To search for \glspl{ANN} that may have more than one hidden-layer we use the grammar of Figure~\ref{fig:multi_layer_grammar}. 

The experimental results are presented in Table~\ref{tab:results_multilayer}. For each metric, the first two rows present the averages of the tests conducted with the grammars of Figures~\ref{fig:one_layer_grammar} and \ref{fig:multi_layer_grammar}, respectively, i.e., the grammars for encoding networks with just one hidden-layer (1 H-L, from Table~\ref{tab:one_layer_results}) or that allow the generation of networks with multiple hidden-layers ($\ge 1$ H-Ls). Additionally, we analyse the \glspl{ANN} that have more than one hidden-layer (last row, $>1$ H-Ls), discarding the evolutionary runs that resulted in networks with just one hidden-layer.

Results of 1 H-L and $\ge 1$ H-Ls are averages of $30$ independent runs; in the $> 1$ H-Ls row results are averages of $21$, $19$, $18$ and $21$ independent runs for the flame, WDBC, ionosphere and sonar datasets, respectively. That is, for the flame, WDBC, ionosphere and sonar datasets in $9$, $11$, $12$ and $9$ runs the best network is composed by only one hidden-layer. For the flame and WDBC we perform runs with $1500$ generations, and for the ionosphere and sonar $3500$ generations are used. The rationale behind the different number of generations is related to the number of features of each problem and the possible progression margin: in flame and WDBC the metrics in train and test already have values close to optimum, thus suggesting that one hidden-layer is likely enough to solve the classification task.

The experimental results show that by using \gls{DSGE} it is possible to evolve effective multi-layered \glspl{ANN}. By comparing the first two rows (1 H-L and $\ge 1$ H-Ls) it is also clear that the results are equivalent, which is confirmed by a statistical analysis (Mann-Whitney, $\alpha$ = 0.05) that shows no statistical differences. Focusing on the comparison between one-hidden-layered \glspl{ANN} and those that have more than one hidden-layer ($>1$ H-Ls) a small, but perceptive difference exists. That difference is statistically significant in some of the training metrics, and in none of the testing metrics. When there is a statistical difference the effect size is always large. Moreover, the difference is larger in the datasets that have more input features and a greater margin for improvement. On the contrary, almost no improvement is observable in the flame dataset, which is expected as the problem only has two features, and thus an \gls{ANN} with one hidden-layer already performs close to optimal. In a nutshell, although there are no statistical differences between the performance of 1 H-L and $\ge 1$ H-Ls runs, when the evolutionary process results in multi-layered \glspl{ANN} the differences begin to emerge. This result shows that \gls{DSGE} is able to cope with the higher dimensionality of the search space associated with the evolution of multi-layered topologies and indicates that in complex datasets, where there is a clear advantage in using deeper topologies, \gls{DSGE} is likely to outperform other grammar-based approaches in train and test.

The complexity of the generated \glspl{ANN} (in terms of the number of used neurons and hidden-layers) is greater when allowing the generation of multi-layered \glspl{ANN}. The difference is even larger if we consider only those networks that have more than one hidden-layer. More neurons means more weights and bias values that have to be evolved, making the evolutionary task more difficult. \glspl{ANN} with fewer layers tend to have less neurons, and consequently are benefited from an evolutionary point of view, as less real values need to be tuned. Thus, in future experiments it is our intention to use the evolutionary approach to evolve the initial set of weights and then apply a fine tuning stage (e.g, backpropagation or resilient backpropagation) during a maximum number of epochs proportional to the number of neurons and/or hidden-layers. As we allow hidden-nodes to connect to nodes in previous layers (including input features) the number of used features is also higher. 

Preliminary experiments concerning the evolution of \glspl{ANN} with more than one output neuron have also been conducted, i.e., instead of using just one output neuron and the sigmoid function as activation function we used two output nodes (one per class) with the softmax activation function. Obtained performances are similar. However, results are not presented due to the lack of space.

\section{Conclusion and Future Work}
\label{sec:conclusions}

In this paper we propose \gls{DSGE}: a new genotypic representation for \gls{SGE}. The gain is two fold: (i) while in previous grammmar-based representations, the genotype encodes the largest allowed sequence, in \gls{DSGE} the genotype grows as needed; and (ii) there is no need to pre-process the grammar in order to expand recursive production rules, as the maximum depth is defined for each sub-tree. Most importantly, \gls{DSGE} solves a limitation of other \gls{GGP} methodologies by allowing the evolution of solutions to dynamic domains, such as \glspl{ANN}, where there is the need to know the number of neurons available in previous layers so that valid individuals are generated.

Results show that \gls{DSGE} is able to evolve the topology and weights of one-hidden-layered \glspl{ANN} that perform statistically better than those evolved using \gls{GE} or \gls{SGE}. Moreover, the results are also better than those provided by hand-crafted \glspl{ANN} finetuned using backpropagation, and than the ones generated using other \gls{GE}-based approaches~\cite{tsoulos2008neural,ahmadizar2015artificial,soltanian2013artificial}. Results concerning the evolution of multi-layered \glspl{ANN} despite not statistical show that the methodology is suitable for the evolution of accurate \glspl{ANN}. Experiments on more, and more difficult datasets are needed to better analyse the evolution of \glspl{ANN} with more than one hidden-layer. However, the ones reported in the current paper had to be used in order to compare \gls{DSGE} with previous approaches.

Future work will focus on the performance of experiments using more complex datasets (e.g., MNIST and CIFAR) and on the generalisation of the allowed layer types, so that it is possible to evolve networks that use, for example, convolution and/or pooling layers.

\section*{Acknowledgments}

\noindent This research is partially funded by: Funda\c{c}\~ao para a Ci\^encia e Tecnologia (FCT), Portugal, under the grant SFRH/BD/114865/2016.

\bibliographystyle{ACM-Reference-Format}
\bibliography{sigproc} 

\end{document}